# From YOLO to VLMs: Advancing Zero-Shot and Few-Shot Detection of Wastewater Treatment Plants Using Satellite Imagery in MENA Region

Akila Premarathna
*Water Futures Data & Analytics*
*International Water Management Institute*
*Colombo, Sri Lanka*
A.Premarathna@cgiar.org

Kanishka Hewageegana
*Water Futures Data & Analytics*
*International Water Management Institute*
*Colombo, Sri Lanka*
K.Hewageegana@cgiar.org

Garcia Andarcia Mariangel
*Water Futures Data & Analytics*
*International Water Management Institute*
*Colombo, Sri Lanka*
M.GarciaAndarcia@cgiar.org

*In regions of the Middle East and North Africa (MENA), there is a high demand for wastewater treatment plants (WWTPs), crucial for sustainable water management. Precise identification of WWTPs from satellite images enables environmental monitoring. Traditional methods like YOLOv8 segmentation require extensive manual labeling. But studies indicate that vision-language models (VLMs) are an efficient alternative to achieving equivalent or superior results through inherent reasoning and annotation. This study presents a structured methodology for VLM comparison, divided into zero-shot and few-shot streams specifically to identify WWTPs. The YOLOv8 was trained on a governmental dataset of 83,566 high-resolution satellite images from Egypt, Saudi Arabia, and UAE: ~85% WWTPs (positives), 15% non-WWTPs (negatives). Evaluated VLMs include LLaMA 3.2 Vision, Qwen 2.5 VL, DeepSeek-VL2, Gemma 3, Gemini, and Pixtral 12B (Mistral), used to identify WWTP components such as circular/rectangular tanks, aeration basins and distinguish confounders via expert prompts producing JSON outputs with confidence and descriptions. The dataset comprises 1,207 validated WWTP locations (198 UAE, 354 KSA, 655 Egypt) and equal non-WWTP sites from field/AI data, as 600m×600m Geo-TIFF images (Zoom 18, EPSG:4326). Zero-shot evaluations on WWTP images showed several VLMs out-performing YOLOv8's true positive rate, with Gemma-3 highest. Results confirm that VLMs, particularly with zero-shot, can replace YOLOv8 for efficient, annotation-free WWTP classification, enabling scalable remote sensing.*

## I. INTRODUCTION

Water scarcity constitutes a fundamental problem in arid and semi-arid regions, where efficient water management strengthens economic growth, public well-being, and environmental harmony [1], [2]. In MENA nations, which have scarce freshwater resources and high urbanization rates, wastewater treatment plants (WWTPs) are significant in water reclamation and reuse that reduces pressure on scarce natural water resources [4], [5]. These plants are utilized to treat domestic, industrial, and agricultural wastewaters for irrigation and industrial reuse [6]. However, it is very difficult to monitor and manage WWTPs across large and remote territories. This statement is accurate in countries such as the United Arab Emirates (UAE), Kingdom of Saudi Arabia (KSA), and Egypt in the Middle East and North Africa (MENA) because they often lack necessary information or data. The main difficulty is keeping track of the environment and making sure the plants follow the rules [8].

High-resolution satellite imaging with the use of remote sensing technologies offers an upper hand for WWTP infrastructure detection and assessment with reduced reliance on time-consuming on-ground surveys [9]. Hence, traditional computer vision algorithms, such as the YOLOv8 object detection model, are effective in discriminating between WWTP structures such as circular/rectangular tanks, aeration basins, and clarifiers from the aerial imagery [11]. YOLOv8 uses convolutional neural networks to achieve real-time processing and high accuracy [12]. Yet, its dependence on large, manually labelled datasets has significant drawbacks, including high labor and time demands, as well as susceptibility to errors in distinguishing WWTPs from visual distractors like industrial tanks or agricultural ponds [13]. These drawbacks are primarily critical in environments with poor resources [14].

**The emergence of VLMs,** which effectively merge computer vision and natural language processing, offers an innovative alternative that addresses the drawbacks experienced with YOLO [15]. With the capability to leverage pre-trained foundation models, VLMs can provide zero-shot and few-shot learning to facilitate task generalization through explanatory prompts without having to rely on enormous, labelled datasets [16]. Their reasoning capability allows semantic interpretation of images and encoded results, i.e., detections in JSON format with confidence scores and descriptive labels, thereby alleviating the annotation burden and promoting efficiency for environmental monitoring remote sensing tasks [14].

This paper compares the efficiency of VLMs for detecting WWTPs from satellite imagery in MENA nations with YOLOv8 segmentation. The author evaluates a set of top-performing VLMs such as; **LLaMA 3.2 Vision, Qwen 2.5 VL, DeepSeek-VL2, Gemma 3 and Gemini,** using a bifurcated method with zero-shot and few-shot test streams. Zero-shot assessment relies on inference-designed prompts by experts, while few-shot assessment taps into a limited set of examples and supervision libraries to calibrate performance. The rest of the paper is organized as follows: Data overview and background, Methodology, Results and Discussion, Conclusion, Acknowledgement, and Reference.

## II. DATA OVERVIEW AND BACKGROUND

This section describes the dataset characteristics and foundational context for detecting WWTPs from satellite imagery, focusing on structural configurations, typologies, and edge cases in the MENA region. The dataset includes 1207 domain expert-validated WWTP sites (198 UAE, 354 KSA, 655 Egypt) and an equal number of non-WWTP sites, imaged as 600m × 600m Geo-TIFF images (Zoom 18, EPSG: 4326).

### A. STRUCTURE OF WWTP

WWTP facilities carry distinctive aerial signatures, including circular or rectangular tanks, aeration basins, clarifiers, and ancillary buildings. The satellite imagery provided highlights of multiple circular tanks (sedimentation or digestion units) and rectangular grids (aeration or filtration areas), often enclosed by fenced perimeters. These elements, in various stages of construction or operation, are the main identifiers for automatic detection. Following section emphasizes the features of WWTPs

#### 1) CONVENTIONAL ACTIVATED SLUDGE SYSTEMS

Include aeration tanks and settling units, visible as rectangular basins and circular tanks.

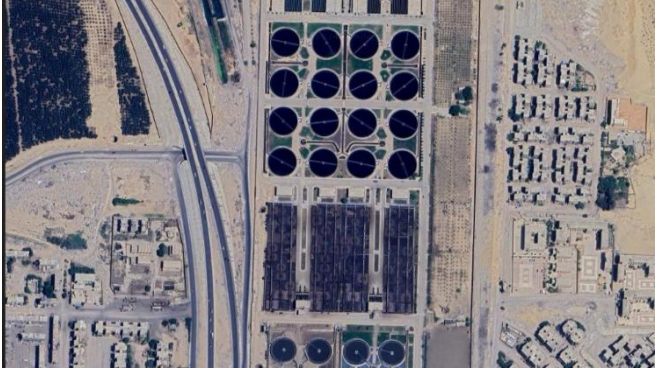

Figure 1: WWTP Location in Egypt

#### 2) TRICKLING FILTERS

Feature rotating arms, identifiable by circular structures with radiating patterns.

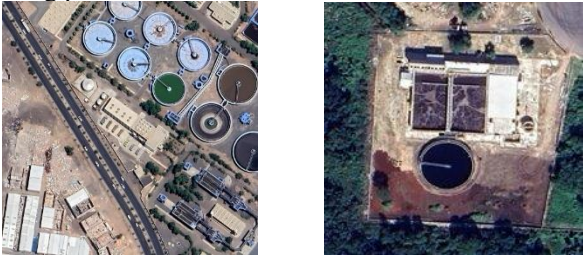

Figure 2: Secondary Clarifier feature of WWTP

### B. EDGE CASES

Tricky lookalikes pose significant challenges to WWTP identification, such as agricultural ponds, industrial silos, or half-built sites that can resemble circular tanks or open areas, making differentiation difficult [25]. The shadows cast by tall structures, as seen in the accompanying image, offer valuable visual cues that can help distinguish these features, enhancing detection accuracy. These variations require models to adapt to dynamic environmental conditions for reliable performance. City growth adds another layer of complexity, as urban sprawl creeping into WWTP edges blurs boundaries, necessitating precise mapping techniques to maintain accuracy in densely developed municipal areas.

Hidden or tiny vegetation plants present additional hurdles by falsely understanding them as circular tanks. With indoor or small-scale WWTPs often covered by roofs or too small to detect, especially without clear shadow indicators [28]. This highlights the importance of improving model sensitivity to capture these subtle cases. Because of these difficulties, we clearly need **smarter tools**. These tools must use cues like **shadows and other visual details** to make detection more reliable in different and complex situations.

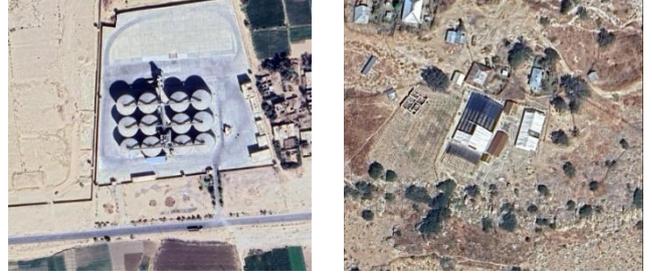

Figure 3: WWTP looks like an edge case

## III. METHODOLOGY

This section outlines the methodological framework to compare a variety of VLMs for detecting WWTPs in satellite imagery from the MENA region. The approach is divided into data acquisition and preparation, model training for YOLOv8, evaluation of VLMs under zero-shot and few-shot paradigms, and performance metrics. All experiments were conducted using Python-based libraries, including **Ultralytics for YOLOv8, Hugging Face Transformers for open-source VLMs, and API integrations for proprietary models like Gemini.**

### A. DATA ACQUISITION & DATASET PREPARATION

To ensure a high-quality and representative dataset, WWTP locations were sourced from multiple channels, followed by image retrieval, annotation, and curation for both training and evaluation purposes. To identify WWTP locations across the MENA region, specifically in Egypt, UAE, and Saudi Arabia, the author compiled data from multiple authoritative sources. This includes official government records, location-based extractions from OpenStreetMap (OSM), and supplementary research conducted by the International Water Management Institute (IWMI).

All collected locations were rigorously validated by domain experts specializing in WWTP infrastructure within the MENA region. These validated sites served as the basis for positive images (those containing actual WWTPs). For negative images (those without WWTPs but visually similar, potentially capable of misleading YOLO or vision-language models), carefully selected analogous locations were also subjected to expert validation to ensure accuracy and relevance.

#### 1) IMAGE METADATA RETRIEVAL

Images were acquired from Google Earth Engine using the Leafmap python package, encompassing a spatial extent of 600 m × 600 m per image. Each image was captured at Zoom level 18 (approximately 0.6 m per pixel), resulting in high-resolution GeoTIFF files projected in the EPSG: 4326 coordinate reference system. RGB bands were prioritized for visual analysis, with preprocessing steps including normalization and cropping to focus on relevant areas. The dataset distribution is summarized in Table 3.1.

TABLE I. DISTRIBUTION OF WWTP AND NON-WWTP LOCATIONS ACROSS COUNTRIES

| Country | WWTP | Non-WWTP |
|---|---|---|
| UAE | 198 | 198 |
| KSA | 354 | 354 |
| Egypt | 655 | 655 |
| Total | 1,207 | 1,207 |

*2) ANNOTATION FOR YOLO TRAINING*

For training the YOLOv8 model, a separate annotated dataset was curated, comprising 83,566 high-resolution satellite images from the same MENA countries. This dataset included georeferenced images with an approximate distribution of 85% positive instances (depicting WWTPs) and 15% negative instances (non-WWTPs). Annotations were performed manually using tools like Roboflow, focusing on instance segmentation masks for WWTP components (tanks, basins) and bounding boxes for whole-site classification.

The dataset was partitioned into a training set of 63,404 images (56,888 positive and 6,516 negatives, from 793 unique locations) and a validation set of 20,110 images (14,321 positive and 5,789 negatives, from 242 unique locations). Negative samples were strategically selected to include confounders like solar panels and water tanks, enhancing the model's robustness. Data augmentation techniques, such as random rotations, flips, and brightness adjustments, were applied during training to increase variability.

*3) EVALUATION DATASET*

The evaluation dataset for both YOLOv8 and VLMs consisted of 2,414 images (1,207 WWTP and 1,207 non-WWTP) described in Image Metadata Retrieval and Annotation for YOLO Training sections. This held-out set ensured unbiased comparison, with no overlap from the YOLO training data. Edge cases, such as partially obscured WWTPs due to cloud cover or urban encroachment, were included to test real-world applicability.

**B. VLM - EVALUATION**

VLMs were evaluated as alternatives to YOLOv8, leveraging their multimodal capabilities for zero-shot and few-shot detection without extensive training. The selected models included open-source (LLaMA 3.2 Vision, Qwen 2.5 VL, DeepSeek-VL2, Gemma 3) and proprietary (Gemini) variants, accessed via Hugging Face or API endpoints.

*1) SELECTED VLMS FOR THE EVALUATION*

The VLMs were chosen for their state-of-the-art performance in vision-language tasks. The selected VLMs are as follows.

TABLE II. CHOSEN VLMs FOR THE STATE-OF-THE-ART PERFORMANCE IN VISION-LANGUAGE TASKS

| Model Name | Developer | Parameter Count | Release Date |
|---|---|---|---|
| LLaMA 3.2 Vision | Meta AI | 11B and 90B (vision models) | September 25, 2024 |
| Qwen 2.5 VL | QwenLM (Alibaba) | 2B, 7B, and 72B variants | August 30, 2024 (initial release, with 72B following) |
| DeepSeek-VL2 | DeepSeek-AI | Tiny: 1B active (3B total), Small: 2.8B active (16B total), Base: 4.5B active | December 13, 2024 (based on available info for small variant) |
| Gemma 3 | Google AI | 1B to 27B (including 270M compact variant) | March 12, 2025 |
| Gemini | Google DeepMind | Not publicly disclosed (proprietary, estimated in trillions for larger versions) | Original: December 2023; Gemini 2.5 Pro: March 25, 2025 |
| Pixtral-12B | Mistral AI (community model) | 12B (multimodal decoder) + 400M (vision encoder) | September 11-17, 2024 |

*2) PROMPT ENGINEERING*

Prompt engineering was pivotal in enabling VLMs to detect WWTPs effectively from satellite imagery, tailored across zero-shot and few-shot paradigms. Prompts were developed by experts in remote sensing and wastewater infrastructure, incorporating detailed step-by-step reasoning to facilitate semantic analysis of visual elements, such as component interconnections and contextual cues specific to MENA arid environments. This structured approach mitigated model hallucinations by including explicit differentiators for confounders and constraints on output format, ensuring consistency.

*3) ZERO-SHOT PROTOCOL*

In the zero-shot configuration, the prompt functioned as a comprehensive definitional guide, requiring the model to "learn" the concept of a WWTP solely from the provided textual instructions without prior visual examples. The prompt explicitly defined a WWTP not merely as isolated tanks, but as a "system of interconnected units." The instructions required the model to scan for specific morphological features:

- Primary Components: Circular tanks (often concrete with central mechanisms), rectangular basins (elongated with a 3:1 to 5:1 ratio), aeration basins (identifiable by surface ripples), and clarifiers.

- Contextual Indicators: The presence of inlet/outlet structures, sludge drying beds (distinguished by color shifts from dark to light), and administrative buildings.
- Differentiation Logic: The prompt included specific instructions to differentiate look-alikes common in the MENA region, such as distinguishing oil/gas facilities (clusters without aeration), agricultural lagoons (irregular shapes), and swimming pools (small, blue features in residential zones).

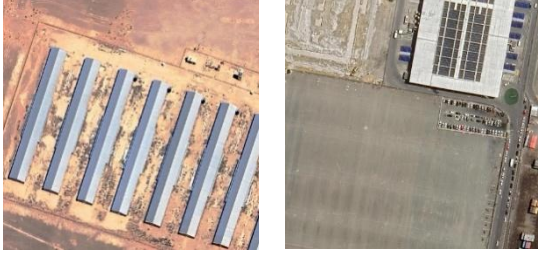

*Figure 4: Examples of a Solar panel and a building*

### 4) FEW-SHOT PROTOCOL AND STRATEGIC SAMPLING.

The few-shot evaluation utilized in-context learning to calibrate the model's performance by embedding specific examples within the prompt. Crucially, the selection of these few-shot exemplars was not random; it followed a targeted error-correction strategy derived from the initial zero-shot results, also validated by a domain expert.

- Addressing Missed Detections (False Negatives): To improve the detection of actual WWTPs that were missed in the zero-shot phase, the few-shot prompt included examples of "hard positives", images of verified WWTPs that lacked clear visual cues or were obscured, which the models had previously failed to identify.
- Addressing False Alarms (False Positives): To reduce the misclassification of non-WWTP sites, the prompt included "hard negatives." These were specific edge cases where the zero-shot models had falsely triggered, such as industrial areas with solar panels, desert areas containing silos, or playgrounds resembling tank structures.

This method allowed the models to learn from their previous boundary errors within the token limit constraints of the inference window.

Output Standardization to ensure automated parsability, both zero-shot and few-shot prompts enforced a strict JSON output schema. The models were instructed to return a binary classification (wwtp: 0 or 1), a count of specific components (circular tanks, rectangular tanks, PSTs), a derived facility type ("Municipal WWTP," "Industrial WWTP"), a confidence score (0.0–1.0), and a brief descriptive string (max 60 characters).

```
{
 "wwtp": 0 or 1,
 "num_circular_tanks": <integer>,
 "num_rectangular_tanks": <integer>,
 "num_pst": <integer>,
 "facility_type": "<string>",
 "confidence": <float>,
 "description": "<string, max 60 characters>"
}
```

## IV. RESULTS AND DISCUSSION

Satellite imagery from the MENA region was used for the evaluation of different models regarding zero-shot detection of WWTPs. The performance metrics were drawn from the resource data, which includes 1207 positive and negative instances. The Table 4.1 presents the confusion matrix for each model, showing the true positives (TP), false negatives (FN), false positives (FP), and true negatives (TN) counts. The Table 4.2 summarizes the precision, recall, and F1-score for each model.

TABLE III. CHOSEN ZERO-SHOT CONFUSION MATRIX

|  | Total = 1207 | | Total = 1207 | |
| --- | --- | --- | --- | --- |
| *Model* | TP | FN | FP | TN |
| **YOLO8** | 990 | 217 | 699 | 509 |
| **DeepSeek-VL2-tiny** | 991 | 216 | 851 | 356 |
| **Gemini 1.5 Flash** | 999 | 208 | 729 | 479 |
| **Gemini 2.0 Flash** | 1074 | 133 | 941 | 267 |
| **Gemini 2.5 Pro** | 1040 | 167 | 712 | 496 |
| **Google Gemma3** | 1171 | 36 | 1186 | 22 |
| **Qwen 2.5 VL 7b** | 958 | 249 | 792 | 416 |
| **llama-3.2-11b-vision** | 1084 | 123 | 1060 | 148 |

TABLE IV. ZERO-SHOT PRECISION, RECALL AND F1-SCORE

| Model | Precision | Recall | F1-Score |
| --- | --- | --- | --- |
| **YOLO8** | 0.5861 | 0.8202 | 0.6837 |
| **DeepSeek-VL2-tiny** | 0.5380 | 0.8210 | 0.6490 |
| **Gemini 1.5 Flash** | 0.5781 | 0.8277 | 0.6808 |
| **Gemini 2.0 Flash** | 0.5330 | 0.8898 | 0.6667 |
| **Gemini 2.5 Pro** | 0.5936 | 0.8616 | 0.7029 |
| **Google Gemma3** | 0.4968 | 0.9702 | 0.6571 |
| **Qwen 2.5 VL 7b** | 0.5474 | 0.7937 | 0.6480 |
| **llama-3.2-11b-vision** | 0.5056 | 0.8981 | 0.6470 |
| **Pixtal 12B** | 0.5029 | 0.8568 | 0.6338 |

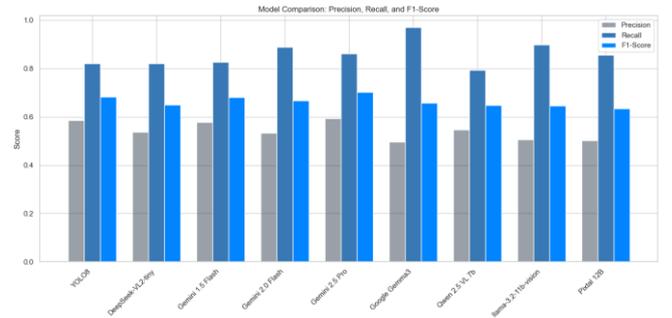

*Figure 5: Zero-shot precision, recall, F1-Score distribution*

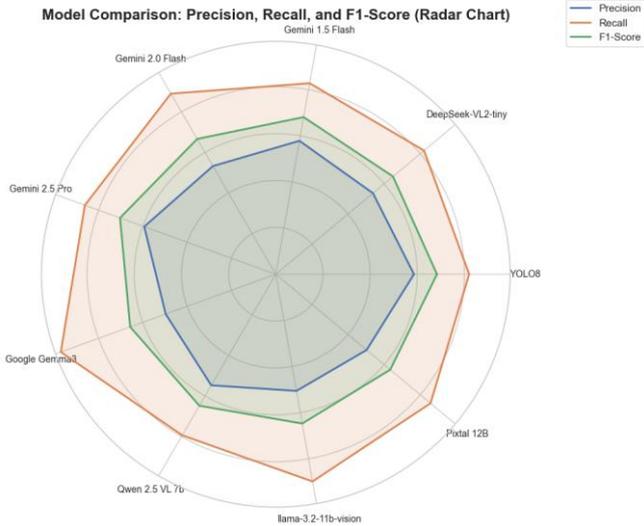

*Figure 6: Spider plots of the Zero-shot precision, recall, and F1-score distribution*

Table 4.3 Few-Shot Confusion Matrix presents the confusion matrix for each model, showing the true positives (TP), false negatives (FN), false positives (FP), and true negatives (TN) counts of the few-shot learning, while Table 4.4 Few-Shot Precision, Recall, and F1 summarizes the precision, recall, and F1-score for each model.

TABLE V.     FEW-SHOT CONFUSION MATRIX

|  | Total = 1207 | | Total = 1208 | |
| --- | --- | --- | --- | --- |
| Model | TP | FN | FP | TN |
| YOLO8 | 990 | 217 | 699 | 509 |
| DeepSeek-VL2-tiny | 918 | 289 | 248 | 960 |
| Gemini 1.5 Flash | - | - | 505 | 703 |
| Gemini 2.0 Flash | 896 | 310 | 665 | 543 |
| Gemini 2.5 Pro | 1028 | 178 | 614 | 594 |
| Google Gemma3 | 1141 | 65 | 190 | 1018 |
| Qwen 2.5 VL 7b | 903 | 303 | 731 | 477 |
| llama-3.2-11b-vision | 1074 | 133 | 960 | 248 |

TABLE VI.     FEW-SHOT PRECISION, RECALL AND F1-SCORE

| Model | Precision | Recall | F1-Score |
| --- | --- | --- | --- |
| YOLO8 | 0.5861 | 0.8202 | 0.6837 |
| DeepSeek-VL2-tiny | 0.7873 | 0.7606 | 0.7737 |
| Gemini 1.5 Flash | - | - | - |
| Gemini 2.0 Flash | 0.5740 | 0.7430 | 0.6476 |
| Gemini 2.5 Pro | 0.6261 | 0.8524 | 0.7219 |
| Google Gemma3 | 0.8572 | 0.9461 | 0.8994 |
| Qwen 2.5 VL 7b | 0.5526 | 0.7488 | 0.6359 |
| llama-3.2-11b-vision | 0.5280 | 0.8898 | 0.6628 |
| Pixtal 12B | 0.5467 | 0.9411 | 0.6917 |

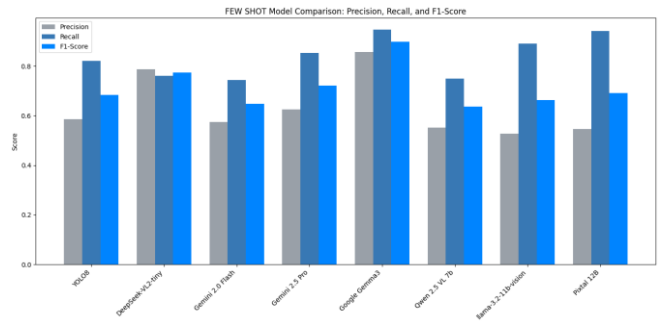

*Figure 7: Few-shot precision, recall, F1-Score distribution*

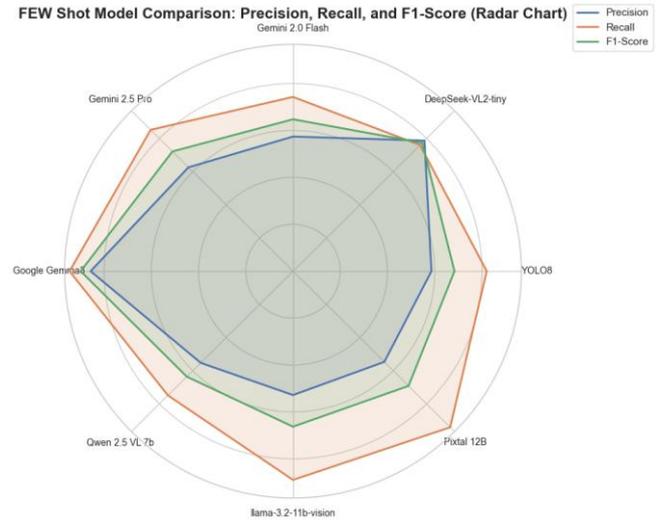

*Figure 8: Spider plots of the Few-shot precision, recall and F1- score distribution*

In summary, the authors compare VLMs against the YOLOv8 benchmark and show very distinct performance characteristics across zero-shot and few-shot paradigms. Additionally, this section analyzes precision, recall, and F1-score metrics to decide on the viability of VLMs as operational alternatives to the traditionally supervised segmentation models for WWTP detection. With this, after comparing the results of zero-shot and few-shot, it highlights that a few-shots outperform the zero-shots with a clear benchmark. The next section will emphasize the comparison in more detail.

*A.  ZERO-SHOT PERFORMANCE*

VLMs manifested a strong bias towards high sensitivity at low specificities. Indeed, Google Gemma 3; LLaMA 3.2 Vision; Pixtral 12B; Gemini 2.0 Flash all obtained the best recall, ranging from 0.89 to 0.97, and often outperforming the recall of baseline YOLOv8 which is 0.8202.

However, this high sensitivity came at the cost of precision. Zero-shot VLMs had a tendency to "over-detect," which resulted in substantial counts of False Positive. For instance, Google Gemma 3 generated 1186 FPs, while Pixtral 12B produced 1041 FPs, yielding the precision values in the range of 0.49-0.58. These are demonstrably lower than YOLOv8's precision of 0.5861. Therefore, while zero-shot VLMs are great at flagging potential candidates, their high noise makes them inapplicable to stand-alone operational monitoring without further filtering mechanisms. YOLOv8 maintained a more balanced performance profile in this comparison.

## B. FEW-SHOT CALIBRATION

Based on the zero-shot prompting results, the author introduced the few-shot prompting acted as a critical stabilizer for VLM performance, drastically reducing false positive rates which indicaes a positive green light. The provision of targeted "hard positive" and "hard negative" examples allowed models to refine their decision boundaries significantly.

- Google Gemma3 drastically increased precision from 0.4968 to 0.8572.
- DeepSeek-VL2-tiny shows a reduction in FPs from 851 (Zero-shot) to 248 (Few-shot), corresponding to a massive precision increase from 0.5380 to 0.7873
- Qwen 2.5 VL reduced FPs from 792 (Zero-shot) to 731 (Few-shot).
- Pixtral 12B, while still noisy, reduced FPs from 1041 (Zero-shot) to below 1000 (Few-shot).

This indicates that even a minimal number of labeled examples can calibrate VLM detection behavior, bringing reliability much closer to, and in some cases surpassing, that of the fully supervised YOLOv8 model. Specifically, Google-Gemma3 and DeepSeek-VL2-tiny's few-shot precision significantly outperformed YOLOv8.

## C. COMPARATIVE ANALYSIS: RECALL AND F1-SCORE

The best-performing few-shot VLMs not only recovered precision but also maintained or exceeded YOLOv8's recall capabilities.

TABLE VII. VLM COMPARATIVE ANALYSIS BASED ON RECALL AND F1-SCORE

| Model | Classification Stream | Recall | F1-Score | Comparison to YOLOv8 F1 ( ≈ 0.6837) |
|---|---|---|---|---|
| DeepSeek-VL2-tiny | Few-Shot | 0.7606 | 0.7737 | Significantly Higher Recall |
| Google Gemma 3 | Few-Shot | **0.9461** | **0.8994** | **Highest F1 Score** |
| Gemini 2.5 Pro | Few-Shot | 0.8524 | 0.7219 | Higher Recall and F1 |
| Llama 3.2 Vision | Few-Shot | 0.8898 | 0.6628 | Higher Recall |
| YOLOv8 (Benchmark) | Baseline | 0.8202 | 0.6837 | Benchmark Performance |

This balance is reflected in the F1-scores, where top-tier few-shot VLMs surpassed the YOLO benchmark (F1 = 0.6837). Google Gemma 3 achieved the highest F1 score across all models at 0.8994, followed by DeepSeek-VL2-tiny at 0.7737 and Gemini 2.5 Pro at 0.7219. These results suggest that few-shot VLMs can offer stronger, balanced detection than traditional methods.

## D. CONFIDENCE SCORE DISTRIBUTION

The following section elaborates on the analysis of the confidence score distributions which reveals the fundamental differences in decision-making behaviors between the supervised YOLOv8 model and the evaluated VLMs.

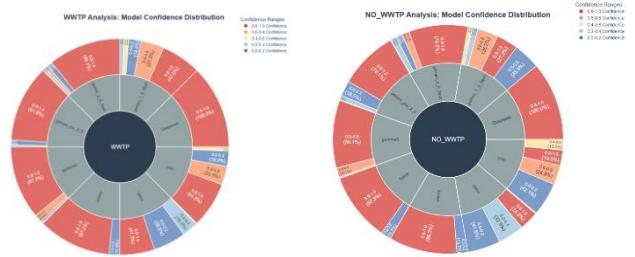

*Figure 9: Confidence interval distribution across YOLO with VLM*

### 1) YOLOV8 CONFIDENCE SCORE

The YOLOv8 model exhibited a highly concentrated confidence distribution, with the vast majority of predictions falling strictly within the 0.8–1.0 and 0.6–0.8 ranges. A negligible fraction of predictions appeared in lower confidence bins. This pattern indicates that YOLOv8 functions as a decisive, low-entropy classifier; once trained, it rarely outputs ambiguous predictions, consistently assigning high certainty to its detections.

### 2) ZERO-SHOT VLMS CONFIDENCE SCORE

Zero-shot VLMs (including Gemini 2.0 Flash, LLaMA 3.2 Vision, Pixtral 12B, and Google Gemma 3) displayed a broad distribution of confidence scores spanning all intervals, including significant clusters in the 0.4–0.6 and even 0.0–0.2 ranges.
This high variability suggests that zero-shot VLMs are more "uncertain" and "exploratory" in their inference. Unlike the rigid pattern recognition of YOLOv8, these models express varying degrees of certainty across different samples. This variance is advantageous as it reflects adaptive reasoning rather than overfitting, indicating that the models are actively weighing visual evidence rather than relying on fixed feature templates.

### 3) FEW-SHOT VLMS CONFIDENCE SCORE

Few-shot prompting induced a measurable shift in VLM confidence distributions toward higher ranges. For models such as DeepSeek-VL2-tiny, Google Gemma 3, and Gemini 2.5 Pro, the data shows:
- A migration of predictions from low-confidence bins (0.0–0.4) into mid-to-high confidence bins.
- A reduction in "noise" at the lower end of the spectrum (0.0–0.2).

This shift confirms that few-shot examples serve to stabilize model behavior.

### E. THE ANNOTATION-PERFORMANCE TRADE-OFF

A critical finding of this study is the scalability advantage offered by VLMs. The YOLOv8 model required a massive annotation effort involving thousands of manually labeled bounding boxes and segmentation masks. In contrast, few-shot VLMs achieved comparable or superior performance (DeepSeek and Gemma 3) using as few as 5 to 30 exemplars. This represents a reduction in annotation workload comparatively, addressing a major bottleneck in remote sensing workflows.

## V. CONCLUSION

This study systematically evaluated the efficacy of Vision-Language Models (VLMs) against the established YOLOv8 object detection framework for identifying Wastewater Treatment Plants (WWTPs) across the arid and semi-arid regions of the Middle East and North Africa (MENA). By leveraging a validated dataset of 1,207 WWTP locations across Egypt, Saudi Arabia, and the UAE, alongside an equal number of confounder sites, we assessed the transition from resource-intensive supervised learning to efficient zero-shot and few-shot paradigms.

The investigation yielded critical insights into the operational readiness of VLMs for remote sensing applications:

- In zero-shot configurations, VLMs such as Google Gemma 3 and LLaMA 3.2 Vision demonstrated exceptional recall (up to 0.97), effectively capturing the vast majority of WWTP infrastructure. However, this high sensitivity was offset by low precision (0.49–0.58) and excessive false positives, confirming that zero-shot VLMs, in their current state, are too noisy to serve as standalone replacements for trained segmentation models like YOLOv8.
- The integration of few-shot prompting, utilizing a strategic selection of "hard positive" and "hard negative" exemplars, fundamentally altered the performance landscape. This approach successfully calibrated the models, drastically reducing false positives (Gemma3 reduced false positives from 1186 to 190) while maintaining high recall.
- Under few-shot conditions, specific VLMs demonstrated superior performance metrics compared to the supervised baseline, like Google Gemma3 achieves 0.8572, and Deepseek tiny VL2 achieves 0.7872. The precision of both surpasses the YOLOv8 benchmark of 0.6837. These results validate the hypothesis that few-shot VLMs can offer a more balanced and accurate detection capability than traditional computer vision models in this domain.
- The most significant operational finding is the dramatic reduction in annotation burden. While YOLOv8 necessitated the manual labeling of over 83,000 images, top-performing VLMs achieved superior accuracy with as few as 5–30 in-context examples. This represents a scalable, cost-efficient pathway for expanding environmental monitoring to new regions without the prohibitive costs of large-scale dataset curation.

In conclusion, while zero-shot VLMs are not yet viable substitutes for traditional methods due to their precision limitations, few-shot VLMs present a transformative alternative. They offer a solid, high-performance solution that matches or even exceeds the reliability of supervised models such as YOLOv8, while requiring only a fraction of the setup effort.

A future enhancement should focus on extending these few-shot methodologies using newer, more advanced models, as well as exploring prompt-based model fine-tuning. Additionally, fine-tuning the model weights based on WWTP structural characteristics may further improve accuracy. Although resource-intensive, such enhancements could push performance to even higher levels.

## VI. ACKNOWLEDGEMENT

The authors gratefully acknowledge the **fund donated by Google.org for the E-ReWater MENA project**, which made this research possible. We extend our sincere gratitude to key contributors for their invaluable support, insights, and dedication throughout the project: **Javier Mateo-Sagasta, Safwat Mahmoud, Joao Diogo Botelheiro, Karthi Matheswaran, and Naga Manohar Velpuri,** whose technical expertise and collaboration were fundamental to the successful development and validation of the models presented in this paper.